\documentclass{article}
\usepackage{xcolor}

 \usepackage[dblblindworkshop, final]{neurips_2025}
\workshoptitle{NeurIPS 2025 Workshop on SPACE in Vision, Language, and Embodied AI (SpaVLE)}

\usepackage{authblk}  
\usepackage[utf8]{inputenc} 
\usepackage[T1]{fontenc}    
\usepackage{hyperref}       
\usepackage{url}            
\usepackage{booktabs}       
\usepackage{amsfonts}       
\usepackage{nicefrac}       
\usepackage{microtype}      
\usepackage{xcolor}         
\usepackage{subcaption} 
\usepackage{lipsum}
\usepackage{wrapfig}
\usepackage{fancyhdr}       
\usepackage{graphicx}       
\usepackage{booktabs}
\usepackage{tabularx} 
\usepackage{amsmath}
\title{TC-LoRA: Temporally Modulated Conditional LoRA for Adaptive Diffusion Control}

\author{
\textbf{Minkyoung Cho$^{1\dagger}$ \quad Ruben Ohana$^{2}$ \quad Christian Jacobsen$^{2}$ \quad Adityan Jothi$^{2}$ \quad Min-Hung~Chen$^{2}$  \quad Z. Morley Mao$^{1}$  \quad Ethem Can$^{2}$}\\
$^{1}$University of Michigan \quad $^{2}$NVIDIA}

\begin{document}

\maketitle
\def\thefootnote{$\dagger$}\footnotetext{Work done during an internship at NVIDIA. Correspondence to: Minkyoung Cho \tt\small <minkycho@umich.edu>}

\begin{abstract}
Current controllable diffusion models typically rely on fixed architectures that modify intermediate activations to inject guidance conditioned on a new modality. This approach uses a static conditioning strategy for a dynamic, multi-stage denoising process, limiting the model's ability to adapt its response as the generation evolves from coarse structure to fine detail.
We introduce TC-LoRA (Temporally Modulated Conditional LoRA), a new paradigm that enables dynamic, context-aware control by conditioning the model's weights directly. Our framework uses a hypernetwork to generate LoRA adapters on-the-fly, tailoring weight modifications for the frozen backbone at each diffusion step based on time and the user's condition. This mechanism enables the model to learn and execute an explicit, adaptive strategy for applying conditional guidance throughout the entire generation process.
Through experiments on various data domains, we demonstrate that this dynamic, parametric control significantly enhances generative fidelity and adherence to spatial conditions compared to static, activation-based methods. TC-LoRA establishes an alternative approach in which the model’s conditioning strategy is modified through a deeper functional adaptation of its weights, allowing control to align with the dynamic demands of the task and generative stage.
\end{abstract}

\section{Introduction}\label{sec:introduction}

The rapid advancement of diffusion models~\cite{ho2020ddpm, song2020scoresde, karras2022edm, liu2022rectifiedflow} has unlocked unprecedented capabilities in image synthesis. A key application is controllable generation for producing synthetic data in critical domains like robotics and autonomous driving~\cite{alhaija2025cosmos-t,agarwal2025cosmos-p}, where real data is expensive to gather and a precise adherence to control signals is crucial. These models invert the traditional data pipeline by generating photorealistic images from precise labels (e.g., depth maps, poses), aiming for perfect label-image correspondence. However, current approaches, such as ControlNet~\cite{zhang2023controlnet}, use a fixed network architecture to process conditioning signals, applying the same conditioning operation throughout the entire denoising process.
This static approach may be suboptimal as the ideal conditioning strategy varies across the diffusion process; for instance, establishing coarse spatial structure is critical in early stages, while refining fine-grained details is the focus of later stages~\cite{balaji2022ediff}. We argue that model weights themselves should be dynamic functions of both diffusion time and the conditioning input, allowing the model to adapt its computational mechanism to the specific needs of each step.

To achieve this, we introduce Temporally Modulated Conditional LoRA (TC-LoRA), a framework based on a hypernetwork that dynamically generates weight adapters conditioned on both time and the control signal. Instead of using static (frozen) weights, TC-LoRA continuously adapts its parameters during inference, representing a paradigm shift towards dynamic weight conditioning. In this work, we take the example of conditioning the diffusion model on depth maps, constraining the generated image to respect the depth. We compare our method with ControlNet-style baselines~\cite{zhang2023controlnet, alhaija2025cosmos-t} and demonstrate its superior adherence to the control signal.

\begin{table}[pt]
\centering
\begin{tabular}{@{}lcc@{}} 
\toprule
 & \textbf{ControlNet-style~\cite{alhaija2025cosmos-t, zhang2023controlnet, chen2024pixart}} 
 & \textbf{TC-LoRA (Ours)} \\
\midrule
Primary Site of Intervention & Activation Space & Weight Space \\
Conditioning Strategy & Static & Dynamic \\
\bottomrule
\end{tabular}
\vspace{1em}
\caption{\textbf{Comparison of Conditioning Mechanisms.} Fundamental differences between ControlNet-style methods~\cite{alhaija2025cosmos-t, zhang2023controlnet, chen2024pixart} and TC-LoRA.}
\label{tab:comparison}
\end{table}

\section{Related Work}\label{sec:prior_work}

\looseness = -1
\textbf{Diffusion Models}~\cite{ho2020ddpm, song2020scoresde, karras2022edm} are generative models that produce high-quality samples by progressively removing noise from an initial random signal. To sample from a target distribution $p(\mathbf{x}_0|\mathbf{c})$, a diffusion model first defines a forward process that incrementally adds Gaussian noise to clean data $\mathbf{x}_0$ over a series of timesteps $t$, producing noisy samples $\mathbf{x}_t$. It then learns a reverse process to remove this noise, conditioned on an input $\mathbf{c}$ (e.g., text, image). Training involves minimizing the difference between the predicted noise and the actual noise added at each step, using a noise prediction network $\epsilon_\theta$: $\mathbb{E}_{\mathbf{x}_0, \mathbf{c}, t \sim \mathcal{T}} \left[ \| \epsilon_\theta(\mathbf{x}_t, t, \mathbf{c}) - \epsilon \|_2^2 \right]$, where $\mathcal{T}$ is the range of timesteps. In modern architectures~(Latent Diffusion Models, LDM), smaller latent representations $\mathbf{z}_t$ are extracted from a pre-trained auto-encoder, so the denoising function $D_\theta$ operating cost is drastically reduced~\cite{rombach2022ldm1, blattmann2023ldm2}.


\textbf{Conditional Control.} 
ControlNet-style models~\cite{alhaija2025cosmos-t, zhang2023controlnet, chen2024pixart} enhance LDMs by incorporating additional spatial conditioning $\mathbf{y}$ (e.g., depth maps, edges) alongside text conditioning $\mathbf{c}$. This is achieved by injecting spatial features from $\mathbf{y}$ into the main denoising network via auxiliary branches, resulting in a function $D_\theta(\mathbf{z}_t, t, \mathbf{c}, \mathbf{y})$. A key characteristic of all these models is that they employ fixed network weights $\theta$ throughout the generation process. Timestep $t$ and conditioning signals $\mathbf{c}$ and $\mathbf{y}$ are treated as inputs processed through static computational pathways, which do not adapt to the varying computational demands across different stages of denoising.

\looseness=-1
\textbf{Low-Rank Adaptation} (LoRA)~\cite{hu2022lora} is a prominent parameter-efficient fine-tuning technique that adapts large pre-trained models by representing weight updates as the product of two low-rank matrices, $\mathbf{B}$ and $\mathbf{A}$. The updated weight is given by $\mathbf{W} = \mathbf{W}_0 + \mathbf{B}\mathbf{A}$, where $\mathbf{W}_0 \in \mathbb{R}^{d \times k}$ are the frozen pre-trained weights, and the trainable matrices $\mathbf{B} \in \mathbb{R}^{d \times r}$ and $\mathbf{A} \in \mathbb{R}^{r \times k}$ have a rank $r \ll \min(d,k)$. In its standard application, LoRA learns a single, fixed set of matrices, resulting in a static adaptation. While some works use hypernetworks to generate these static adapters based on conditioning inputs~\cite{ha2016hypernetworks, charakorn2025t2l}, recent efforts have focused on making adaptations dynamic. For instance, T-LoRA~\cite{soboleva2025tlora} and Time-Varying LoRA~\cite{zhuang2024timevaryinglora} introduce a time-dependent scaling factor to modulate the magnitude of the LoRA update over a process. However, in these approaches, the underlying low-rank adapters ($\mathbf{B}$ and $\mathbf{A}$) remain functionally static, with only their magnitude being scaled.

\section{TC-LoRA: Temporally Modulated Conditional LoRA}
\label{sec:method}

\paragraph{Motivation for Dynamic Weight Adaptation.}
Current controllable diffusion models utilize fixed network weights, treating timestep and conditioning information as inputs. We argue this approach is inherently less expressive than directly modifying the weights themselves. A model with fixed weights can only modulate its output through learned nonlinearities within a static computational structure. We demonstrate in Appendix~\ref{sec:appendix_distinction} why adding to the activations of the neural network -- i.e. the case of ControlNet -- is not mathematically equivalent to a modification of the weights.

In contrast, a model that dynamically adapts its weights can fundamentally alter its computational mechanisms, enabling qualitatively different processing strategies for each stage of generation. This is analogous to the distinction between modulating the inputs to a fixed function, $f_\theta(x, t, y)$, and changing the function itself, $f_{\theta(t,y)}(x, t, y)$. Research in dynamic networks and neural architecture search supports the hypothesis that adaptive network structures offer superior performance over fixed architectures~\cite{NAS-White-2023, dynamic-networks-Han-2021}. We therefore propose to replace the standard fixed denoising function $D_\theta(\mathbf{z}_t, t, \mathbf{c}, \mathbf{y})$ with a dynamic counterpart, $D_{\theta(t,\mathbf{y})}(\mathbf{z}_t, t, \mathbf{c}, \mathbf{y})$, where the weights $\theta(t, \mathbf{y})$ adapt throughout generation. This raises the central question of how to efficiently parameterize and learn such dynamic weights while leveraging powerful pre-trained models.

To realize dynamic weight adaptation efficiently, we introduce TC-LoRA (Temporally Modulated Conditional LoRA), which combines the parameter efficiency of LoRA with the dynamic capabilities of hypernetworks.
Our approach generates time and condition-specific LoRA weights that are injected into the pre-trained diffusion model, enabling adaptive processing while preserving the base model's learned representations.

\paragraph{TC-LoRA Methodology.}
Given a pre-trained denoising function $D_\theta$ with weights $\theta = \{W_0, \dots, W_N\}$, our method, TC-LoRA, modifies specific weight groups $W_i \in \mathbb{R}^{d\times k}$ by adding dynamic low-rank adaptations. During both training and inference, the weights are updated as a function of the diffusion timestep $t$ and a conditioning signal $\mathbf{y}$:

\begin{equation}
\mathbf{W}'_i = \mathbf{W}_i + \mathbf{B}(i,t, \mathbf{y})\mathbf{A}(i,t, \mathbf{y}),
\end{equation}

where $\mathbf{B}(i, t, \mathbf{y}) \in \mathbb{R}^{d \times r}$ and $\mathbf{A}(i, t, \mathbf{y}) \in \mathbb{R}^{r \times k}$ are low-rank matrices ($r \ll \min(d,k)$) that are not trained directly. Instead, they are generated on-the-fly by a single hypernetwork $H_\phi$, shared across layers. This hypernetwork takes the weight group index $i$, timestep $t$, and conditioning $\mathbf{y}$ as input to produce the complete set of adapter parameters $\{\mathbf{A}(i, t, \mathbf{y}), \mathbf{B}(i, t, \mathbf{y})\}$ for that specific context.

\begin{figure}
  \centering
  \includegraphics[width=1\linewidth]{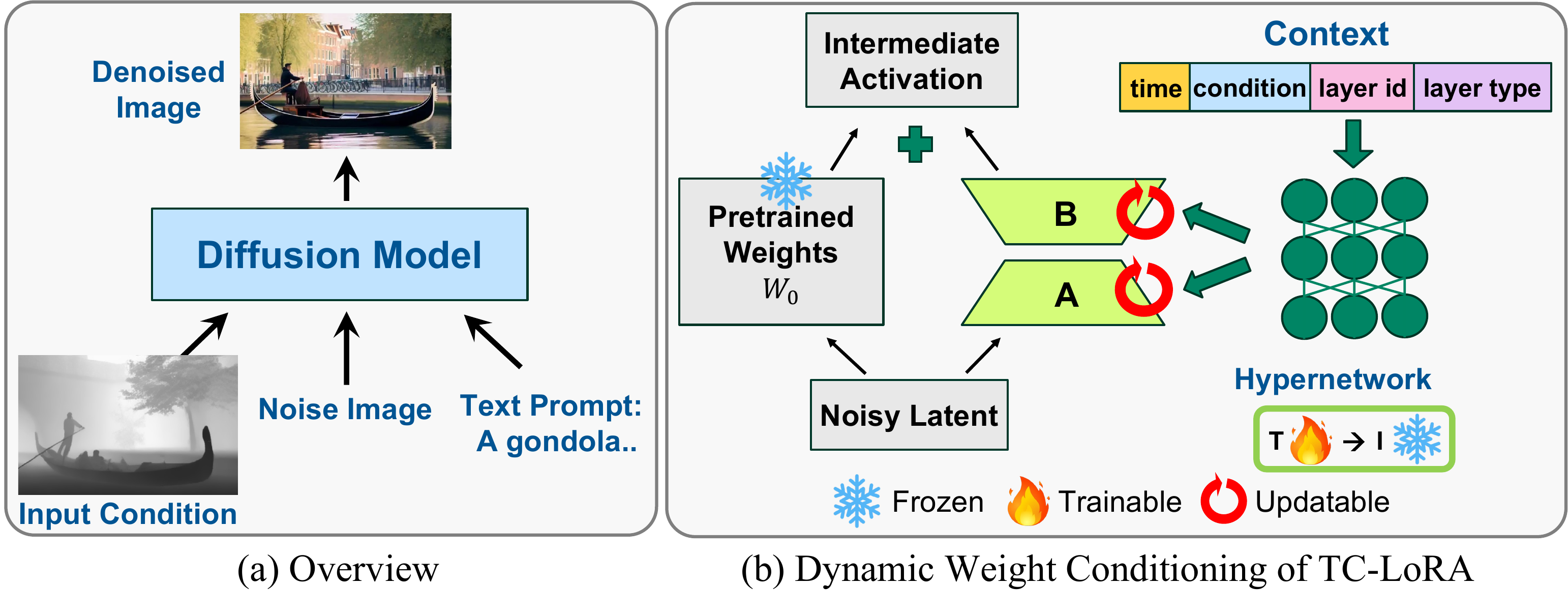}
  \caption{\textbf{The TC-LoRA framework for dynamic, context-aware conditional control.} 
    \textbf{(a)} Our method is formulated within a standard conditional diffusion framework, in which a text prompt together with an input condition (e.g., a depth map or an edge sketch) guides the generation process from an initial noise image.
    \textbf{(b)} We introduce a dynamic weight conditioning mechanism where a hypernetwork---the only trainable component---generates the LoRA adapter weights~($\mathbf{B}$ or $\mathbf{A}$) on-the-fly. This dynamic weight update is then used to modify the behavior of the frozen pretrained weights ($W_0$) of the base model, allowing for a state-aware adaptation at each step.}
    
  \label{fig:main_arch}
\end{figure}

As illustrated in Figure~\ref{fig:main_arch}, by modifying the processing weights themselves, this approach reshapes dynamically the function defined by the diffusion model, allowing the model's generative capacity to be more effectively constrained by the conditioning input, rather than merely influencing activations within fixed computational functions. Further details on the hypernetwork structure and training objective are provided in Appendices~\ref{appendix:A} and~\ref{appendix:B}, respectively.

\begin{table}[tp]
    \centering
    \small 
    \setlength{\tabcolsep}{8pt} 
    \renewcommand{\arraystretch}{1} 
    \begin{tabular}{lccccc}
        \toprule
        & \multicolumn{2}{c}{\textbf{OpenImages~\cite{kuznetsova2020openimages}}} & \multicolumn{2}{c}{\textbf{TransferBench~\cite{alhaija2025cosmos-t}}} \\
        \cmidrule(lr){2-3} \cmidrule(lr){4-5}
        \textbf{Metric} & \textbf{ControlNet~\cite{zhang2023controlnet}} & \textbf{TC-LoRA} & \textbf{ControlNet~\cite{zhang2023controlnet}} & \textbf{TC-LoRA} \\
        \midrule
        NMSE~\cite{ohana2024well}  ($\downarrow$)   & 0.7433 & \textbf{0.7354} & 0.5130 & \textbf{0.4529} \\
        si-MSE~\cite{eigen2014sirmse} ($\downarrow$) & 1.5633 & \textbf{1.0557} & 1.7080 & \textbf{1.6499} \\
        \bottomrule
    \end{tabular}
    \vspace{1em}
    \caption{\textbf{Quantitative comparison on OpenImages benchmark~\cite{kuznetsova2020openimages} and TransferBench~\cite{alhaija2025cosmos-t}}. Lower values are better. Best results for each benchmark are in bold.}
    \label{tab:eval_table}
\end{table}

\begin{figure}[tp]
    
    \includegraphics[width=\linewidth]{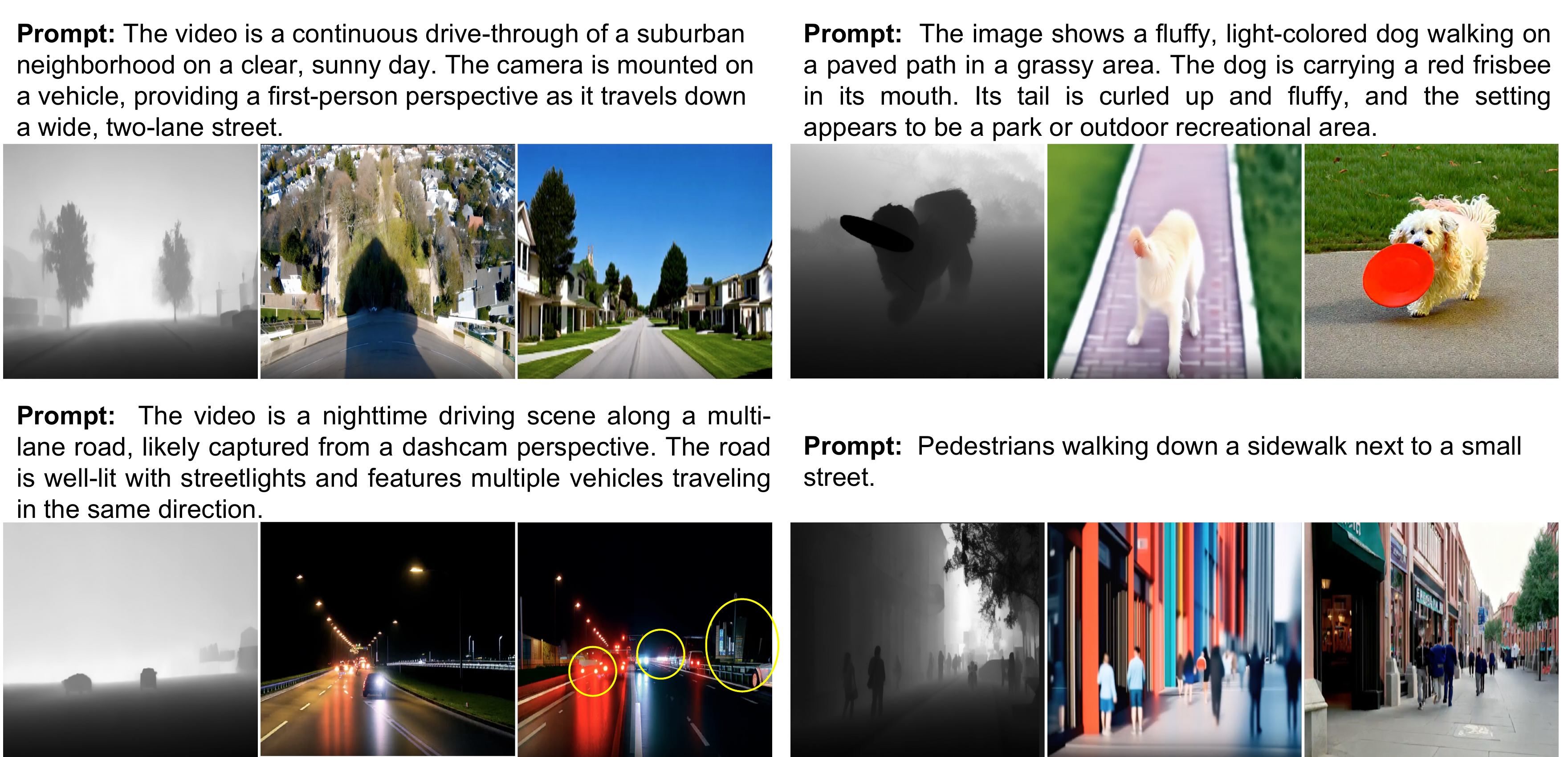}
    \caption{\textbf{Qualitative Comparison:} Each example set comprises a text prompt (top), a depth condition (left image), and the corresponding generation results obtained using ControlNet (middle image) and TC-LoRA conditioning (right image). Specifically, we compare Cosmos-Transfer1~\cite{alhaija2025cosmos-t} against Cosmos-Predict1 + TC-LoRA. Overall, TC-LoRA exhibits improved fidelity in adhering to the spatial condition.}
    \label{fig:eval_transfer1}
\end{figure}

\section{Experimental Results}\label{sec:experiments}

We evaluate the conditioning performance of TC-LoRA on the base model Cosmos-Predict1~\cite{agarwal2025cosmos-p}), and its ability to align image generation with provided depth maps, while noting that it can be generalized to other modalities e.g. edge maps, normal maps, bounding boxes. The TC-LoRA adapters are trained exclusively on the MS-COCO dataset~\cite{lin2014coco} for 3 days using 8 NVIDIA H100 96GB GPUs with a batch size of 4. We then evaluate on two distinct benchmarks to test generalization: a custom-curated OpenImages Benchmark~\cite{kuznetsova2020openimages} with diverse natural, urban, and indoor scenes, and TransferBench~\cite{alhaija2025cosmos-t}, which contains various scenes from robotics~\cite{bu2025agibot}, driving~\cite{yang2024opendv}, and egocentric domains~\cite{grauman2024egoexo}. 

For a fair comparison, our method is benchmarked against Cosmos-Transfer1~\cite{alhaija2025cosmos-t}, which augments the same frozen base model with a ControlNet-style architecture for activation-based conditioning. Alignment fidelity is measured using scale-invariant MSE (si-MSE)~\cite{eigen2014sirmse} and Normalized MSE (NMSE)~\cite{ohana2024well}, where lower values indicate better performance. Notably, TC-LoRA achieves this control with significantly fewer trainable parameters than the baseline (251M vs. 900M). Further details on the benchmarks, metrics, and baseline are available in Appendix~\ref{appendix:exp_details}.

\paragraph{Results and Analysis.}

Our results validate that TC-LoRA's dynamic, weight-based adaptation significantly enhances the model's adherence to spatial conditions. On the OpenImages Benchmark, this is demonstrated by a markedly lower si-MSE, indicating a high degree of fidelity to the depth condition (Table~\ref{tab:eval_table}). This effective generalization, despite being trained only on MS-COCO, continues on TransferBench, where TC-LoRA shows consistent improvement, reducing the NMSE by 11.7\% and the si-MSE by 3.4\% compared to the baseline.

These quantitative improvements are visually reflected in our qualitative analysis (Figure~\ref{fig:eval_transfer1}), where TC-LoRA shows a strong capability for preserving fine-grained details—such as a dog's specific pose or the silhouettes of pedestrians—that the baseline struggles to capture. Furthermore, we illustrate the learning progression in Appendix Figure~\ref{fig:eval_effect}, which shows the model's generation output evolving from an unconditioned state to a high-fidelity, structurally consistent image after 150k training iterations. Together, these results underscore the effectiveness of our dynamic weight conditioning approach for achieving a more precise and robust alignment with user-provided guidance, with further analysis provided in Appendix~\ref{appendix:C.4}.

\textbf{Conclusion and Future Work.} In this work, we introduced TC-LoRA, a new framework that provides dynamic spatial control in diffusion models by regenerating LoRA weights at each denoising step. This method allows the model to learn an adaptive strategy, tailoring its parameters to the evolving needs of the generation process—from establishing coarse structures to refining fine details. By enabling a deeper form of functional adaptation, TC-LoRA significantly enhances the steerability and efficiency of generative foundation models.

A promising future direction is extending this framework to text-to-video generation. The key challenge is to maintain temporal consistency across frames while adhering to per-frame spatial conditions. We propose adapting the hypernetwork to process features from previous frames, enabling TC-LoRA to learn a balance between conditional accuracy and smooth temporal transitions, representing a significant step towards more coherent and controllable video synthesis.

\bibliographystyle{unsrt}
\bibliography{references}

\begin{thebibliography}{10}

\bibitem{ho2020ddpm}
Jonathan Ho, Ajay Jain, and Pieter Abbeel.
\newblock Denoising diffusion probabilistic models.
\newblock {\em Advances in neural information processing systems}, 33:6840--6851, 2020.

\bibitem{song2020scoresde}
Yang Song, Jascha Sohl-Dickstein, Diederik~P Kingma, Abhishek Kumar, Stefano Ermon, and Ben Poole.
\newblock Score-based generative modeling through stochastic differential equations.
\newblock {\em arXiv preprint arXiv:2011.13456}, 2020.

\bibitem{karras2022edm}
Tero Karras, Miika Aittala, Timo Aila, and Samuli Laine.
\newblock Elucidating the design space of diffusion-based generative models.
\newblock {\em Advances in neural information processing systems}, 35:26565--26577, 2022.

\bibitem{liu2022rectifiedflow}
Qiang Liu.
\newblock Rectified flow: A marginal preserving approach to optimal transport.
\newblock {\em arXiv preprint arXiv:2209.14577}, 2022.

\bibitem{alhaija2025cosmos-t}
Hassan~Abu Alhaija, Jose Alvarez, Maciej Bala, Tiffany Cai, Tianshi Cao, Liz Cha, Joshua Chen, Mike Chen, Francesco Ferroni, Sanja Fidler, et~al.
\newblock Cosmos-transfer1: Conditional world generation with adaptive multimodal control.
\newblock {\em arXiv preprint arXiv:2503.14492}, 2025.

\bibitem{agarwal2025cosmos-p}
Niket Agarwal, Arslan Ali, Maciej Bala, Yogesh Balaji, Erik Barker, Tiffany Cai, Prithvijit Chattopadhyay, Yongxin Chen, Yin Cui, Yifan Ding, et~al.
\newblock Cosmos world foundation model platform for physical ai.
\newblock {\em arXiv preprint arXiv:2501.03575}, 2025.

\bibitem{zhang2023controlnet}
Lvmin Zhang, Anyi Rao, and Maneesh Agrawala.
\newblock Adding conditional control to text-to-image diffusion models.
\newblock In {\em Proceedings of the IEEE/CVF international conference on computer vision}, pages 3836--3847, 2023.

\bibitem{balaji2022ediff}
Yogesh Balaji, Seungjun Nah, Xun Huang, Arash Vahdat, Jiaming Song, Qinsheng Zhang, Karsten Kreis, Miika Aittala, Timo Aila, Samuli Laine, et~al.
\newblock ediff-i: Text-to-image diffusion models with an ensemble of expert denoisers.
\newblock {\em arXiv preprint arXiv:2211.01324}, 2022.

\bibitem{chen2024pixart}
Junsong Chen, Yue Wu, Simian Luo, Enze Xie, Sayak Paul, Ping Luo, Hang Zhao, and Zhenguo Li.
\newblock Pixart-$\{$$\backslash$delta$\}$: Fast and controllable image generation with latent consistency models.
\newblock {\em arXiv preprint arXiv:2401.05252}, 2024.

\bibitem{rombach2022ldm1}
Robin Rombach, Andreas Blattmann, Dominik Lorenz, Patrick Esser, and Bj{\"o}rn Ommer.
\newblock High-resolution image synthesis with latent diffusion models.
\newblock In {\em Proceedings of the IEEE/CVF conference on computer vision and pattern recognition}, pages 10684--10695, 2022.

\bibitem{blattmann2023ldm2}
Andreas Blattmann, Robin Rombach, Huan Ling, Tim Dockhorn, Seung~Wook Kim, Sanja Fidler, and Karsten Kreis.
\newblock Align your latents: High-resolution video synthesis with latent diffusion models.
\newblock In {\em Proceedings of the IEEE/CVF conference on computer vision and pattern recognition}, pages 22563--22575, 2023.

\bibitem{hu2022lora}
Edward~J Hu, Yelong Shen, Phillip Wallis, Zeyuan Allen-Zhu, Yuanzhi Li, Shean Wang, Lu~Wang, Weizhu Chen, et~al.
\newblock Lora: Low-rank adaptation of large language models.
\newblock {\em ICLR}, 1(2):3, 2022.

\bibitem{ha2016hypernetworks}
David Ha, Andrew Dai, and Quoc~V Le.
\newblock Hypernetworks.
\newblock {\em arXiv preprint arXiv:1609.09106}, 2016.

\bibitem{charakorn2025t2l}
Rujikorn Charakorn, Edoardo Cetin, Yujin Tang, and Robert~Tjarko Lange.
\newblock Text-to-lora: Instant transformer adaption.
\newblock {\em arXiv preprint arXiv:2506.06105}, 2025.

\bibitem{soboleva2025tlora}
Vera Soboleva, Aibek Alanov, Andrey Kuznetsov, and Konstantin Sobolev.
\newblock T-lora: Single image diffusion model customization without overfitting.
\newblock {\em arXiv preprint arXiv:2507.05964}, 2025.

\bibitem{zhuang2024timevaryinglora}
Zhan Zhuang, Yulong Zhang, Xuehao Wang, Jiangang Lu, Ying Wei, and Yu~Zhang.
\newblock Time-varying lora: Towards effective cross-domain fine-tuning of diffusion models.
\newblock {\em Advances in Neural Information Processing Systems}, 37:73920--73951, 2024.

\bibitem{NAS-White-2023}
C.~White.
\newblock Neural architecture search.
\newblock {\em Foundations and Trends in Machine Learning}, 2023.

\bibitem{dynamic-networks-Han-2021}
Y.~Han.
\newblock Dynamic neural networks: A survey.
\newblock {\em IEEE Transactions on Pattern Analysis and Machine Intelligence}, 2021.

\bibitem{kuznetsova2020openimages}
Alina Kuznetsova, Hassan Rom, Neil Alldrin, Jasper Uijlings, Ivan Krasin, Jordi Pont-Tuset, Shahab Kamali, Stefan Popov, Matteo Malloci, Alexander Kolesnikov, et~al.
\newblock The open images dataset v4: Unified image classification, object detection, and visual relationship detection at scale.
\newblock {\em International journal of computer vision}, 128(7):1956--1981, 2020.

\bibitem{ohana2024well}
Ruben Ohana, Michael McCabe, Lucas Meyer, Rudy Morel, Fruzsina Agocs, Miguel Beneitez, Marsha Berger, Blakesly Burkhart, Stuart Dalziel, Drummond Fielding, et~al.
\newblock The well: a large-scale collection of diverse physics simulations for machine learning.
\newblock {\em Advances in Neural Information Processing Systems}, 37:44989--45037, 2024.

\bibitem{eigen2014sirmse}
David Eigen, Christian Puhrsch, and Rob Fergus.
\newblock Depth map prediction from a single image using a multi-scale deep network.
\newblock {\em Advances in neural information processing systems}, 27, 2014.

\bibitem{lin2014coco}
Tsung-Yi Lin, Michael Maire, Serge Belongie, James Hays, Pietro Perona, Deva Ramanan, Piotr Doll{\'a}r, and C~Lawrence Zitnick.
\newblock Microsoft coco: Common objects in context.
\newblock In {\em European conference on computer vision}, pages 740--755. Springer, 2014.

\bibitem{bu2025agibot}
Qingwen Bu, Jisong Cai, Li~Chen, Xiuqi Cui, Yan Ding, Siyuan Feng, Shenyuan Gao, Xindong He, Xuan Hu, Xu~Huang, et~al.
\newblock Agibot world colosseo: A large-scale manipulation platform for scalable and intelligent embodied systems.
\newblock {\em arXiv preprint arXiv:2503.06669}, 2025.

\bibitem{yang2024opendv}
Jiazhi Yang, Shenyuan Gao, Yihang Qiu, Li~Chen, Tianyu Li, Bo~Dai, Kashyap Chitta, Penghao Wu, Jia Zeng, Ping Luo, et~al.
\newblock Generalized predictive model for autonomous driving.
\newblock In {\em Proceedings of the IEEE/CVF Conference on Computer Vision and Pattern Recognition}, pages 14662--14672, 2024.

\bibitem{grauman2024egoexo}
Kristen Grauman, Andrew Westbury, Lorenzo Torresani, Kris Kitani, Jitendra Malik, Triantafyllos Afouras, Kumar Ashutosh, Vijay Baiyya, Siddhant Bansal, Bikram Boote, et~al.
\newblock Ego-exo4d: Understanding skilled human activity from first-and third-person perspectives.
\newblock In {\em Proceedings of the IEEE/CVF Conference on Computer Vision and Pattern Recognition}, pages 19383--19400, 2024.

\bibitem{cosmos-tokenizer}
Fitsum Reda, Jinwei Gu, Xian Liu, Songwei Ge, Ting-Chun Wang, Haoxiang Wang, and Ming-Yu Liu.
\newblock Nvidia blog: https://research.nvidia.com/labs/dir/cosmos-tokenizer/.

\bibitem{sun2024sinu2}
Chuanhao Sun, Zhihang Yuan, Kai Xu, Luo Mai, N~Siddharth, Shuo Chen, and Mahesh~K Marina.
\newblock Learning high-frequency functions made easy with sinusoidal positional encoding.
\newblock {\em arXiv preprint arXiv:2407.09370}, 2024.

\bibitem{radford2021clip}
Alec Radford, Jong~Wook Kim, Chris Hallacy, Aditya Ramesh, Gabriel Goh, Sandhini Agarwal, Girish Sastry, Amanda Askell, Pamela Mishkin, Jack Clark, et~al.
\newblock Learning transferable visual models from natural language supervision.
\newblock In {\em International conference on machine learning}, pages 8748--8763. PmLR, 2021.

\bibitem{xiao2024florence}
Bin Xiao, Haiping Wu, Weijian Xu, Xiyang Dai, Houdong Hu, Yumao Lu, Michael Zeng, Ce~Liu, and Lu~Yuan.
\newblock Florence-2: Advancing a unified representation for a variety of vision tasks.
\newblock In {\em Proceedings of the IEEE/CVF Conference on Computer Vision and Pattern Recognition}, pages 4818--4829, 2024.

\bibitem{ke2025marigold}
Bingxin Ke, Kevin Qu, Tianfu Wang, Nando Metzger, Shengyu Huang, Bo~Li, Anton Obukhov, and Konrad Schindler.
\newblock Marigold: Affordable adaptation of diffusion-based image generators for image analysis.
\newblock {\em arXiv preprint arXiv:2505.09358}, 2025.

\end{thebibliography}

\appendix

\section*{Appendix}
\section{Hypernetwork Architecture}
\label{appendix:A}

\begin{wrapfigure}{r}{0.35\textwidth}
    \centering
    \vspace{-5pt}
    \includegraphics[width=\linewidth]{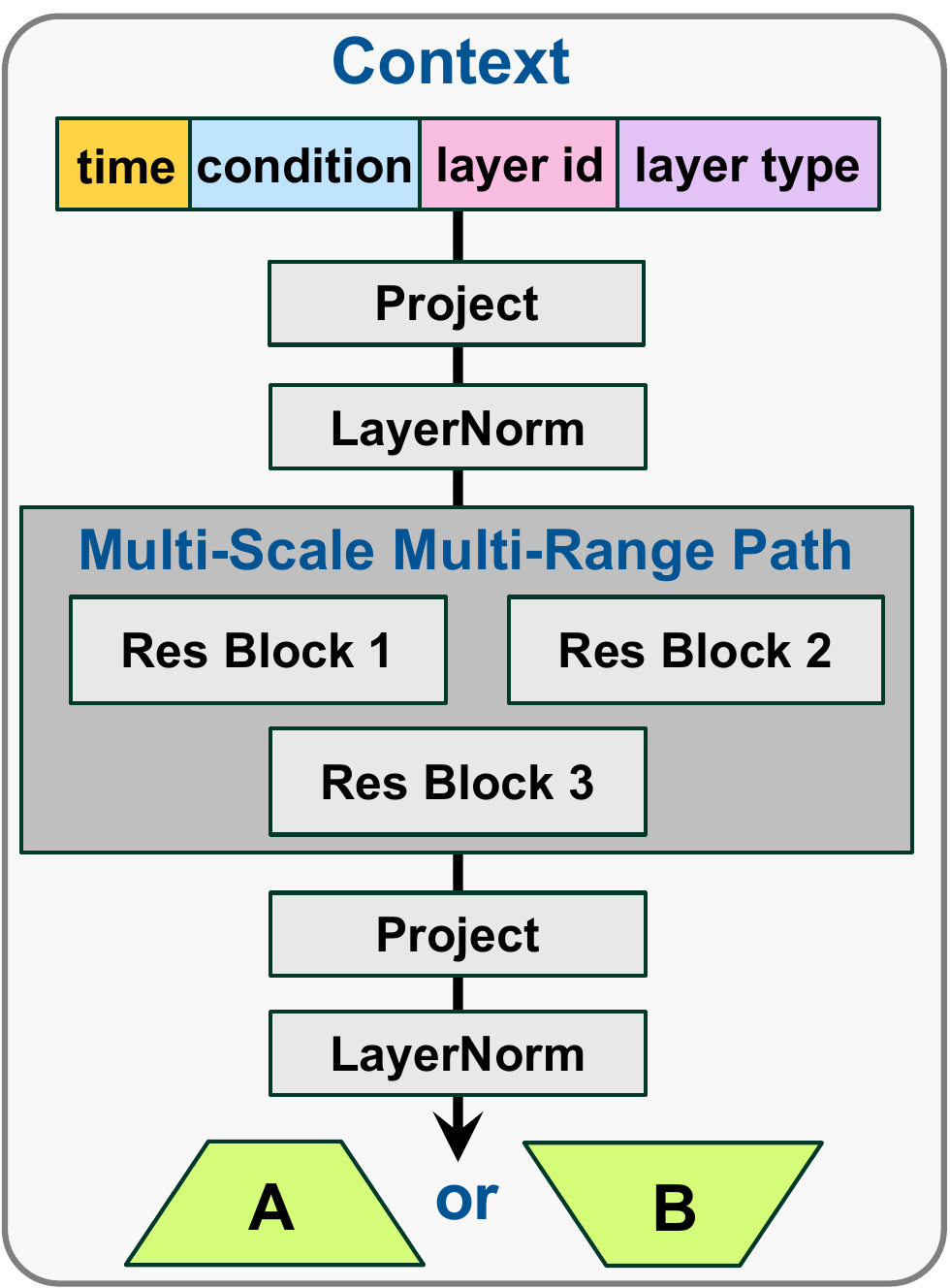}
    \caption{Hypernetwork Design}
    \label{fig:hypernetwork_details}
\end{wrapfigure}


Our architecture employs a single, parameter-efficient hypernetwork, $H_{\phi}$, to generate the low-rank adaptation matrices, $\mathbf{B}$ and $\mathbf{A}$, for all targeted layers within the foundation model. To produce specialized parameters for different contexts, the hypernetwork is conditioned on a context vector that encapsulates four distinct types of information: the diffusion timestep, the user's input condition, the target layer's identity, and the layer's type as shown in Figure~\ref{fig:hypernetwork_details}.

The input condition is first processed into the model's native latent space using the pre-trained autoencoder from the base model~\cite{agarwal2025cosmos-p, cosmos-tokenizer}. This latent representation is then passed through a dedicated 3-layer MLP to produce a fixed-size 1024-dimensional condition embedding. The diffusion timestep is handled by a standard sinusoidal embedding~\cite{agarwal2025cosmos-p, sun2024sinu2}, which yields a 64-dimensional time embedding. Finally, a specialized Layer ID encoder with residual connections maps the target layer's structural properties—a concatenated vector of its depth (ID) and type (e.g., self-attention vs. cross-attention, and query vs. key vs. value)—to a dense 128-dimensional embedding. These three embeddings are concatenated to form the final context vector that is fed into the hypernetwork.

As detailed in Figure~\ref{fig:hypernetwork_details}, the internal architecture of the hypernetwork processes this fused context vector through an input stage followed by a series of multi-scale residual blocks. A key feature of our design is the use of multi-range skip connections, which project features from early and intermediate stages and add them to the final output projection. This allows the hypernetwork to effectively combine low-level and high-level contextual information when generating the adapter parameters. To ensure training stability and mirror standard LoRA initialization, the final layer of the hypernetwork that generates the $\mathbf{B}$ matrix is explicitly zero-initialized. This ensures that our adaptation has no initial effect at the start of training, making the model's output identical to that of the base model. Within the DiT-based foundation model, these dynamic adapters are attached to the linear projection layers in all self-attention and cross-attention blocks, as illustrated in Figures~\ref{fig:main_arch}.

\section{Training Objective}
\label{appendix:B}
TC-LoRA is trained end-to-end with the standard diffusion objective, where the hypernetwork learns to predict appropriate adaptations for each timestep-condition pair:

\begin{equation}
\mathcal{L} = \mathbb{E}_{\mathbf{z}_0, \mathbf{c}, \mathbf{y}, t, \boldsymbol{\epsilon}} \left[ \| \boldsymbol{\epsilon} - D_{\theta + H_\phi(t,\mathbf{y})}(\mathbf{z}_t, t, \mathbf{c}, \mathbf{y}) \|^2 \right]
\end{equation}

Here, the loss $\mathcal{L}$ is the expected squared error between the sampled noise $\boldsymbol{\epsilon}$ and the prediction from the denoising model $D$, which operates on the noisy latent $\mathbf{z}_t$ (derived from clean latent $\mathbf{z}_0$); the model's frozen base weights $\theta$ are dynamically adapted by our trainable hypernetwork $H_\phi$, which itself is conditioned on the diffusion timestep $t$ alongside text conditioning $\mathbf{c}$ and spatial conditioning $\mathbf{y}$ (e.g., depth maps, edges).
This formulation allows the hypernetwork to learn optimal weight adaptations that minimize reconstruction error across all timestep-condition combinations, naturally discovering adaptive processing strategies that improve controllable generation.

\section{Additional Experimental Details and Results}
\label{appendix:exp_details}

\subsection{Benchmark and Dataset Details}
\noindent\textbf{Training Dataset.} All TC-LoRA adapters were trained on the MS-COCO dataset~\cite{lin2014coco}, which contains 120k images of natural and urban scenes. No training was performed on the evaluation benchmark datasets.

\noindent\textbf{OpenImages Benchmark.} We constructed this benchmark with 600 samples for topical balance and visual diversity. Using captions, we first classified samples from the OpenImages dataset~\cite{kuznetsova2020openimages} into three topics: nature, urban, and indoor, using a zero-shot classifier. Within each topic, we used CLIP embeddings~\cite{radford2021clip} and a greedy max-min algorithm for diversity sampling to select visually dissimilar images. This resulted in 200 diverse samples per topic. Text prompts were generated for each image using Florence 2-L~\cite{xiao2024florence}.

\noindent\textbf{TransferBench.} We use the TransferBench benchmark introduced in Cosmos-Transfer1~\cite{alhaija2025cosmos-t}. It consists of 600 examples, which have been slightly modified for our target task of single-frame generation, although the benchmark was originally for text-to-video generation. For this study, we focus on generating the first frame of each example video. The topics are distributed across three challenging, out-of-distribution domains: robotic arm operations (AgiBot World~\cite{bu2025agibot}), driving (OpenDV~\cite{yang2024opendv}), and everyday life scenes (Ego-Exo-4D~\cite{grauman2024egoexo}).

\subsection{Evaluation Metric Details}
To quantify the alignment between the input depth map and the generated image, we first extract a depth map from the generated image using Marigold~\cite{ke2025marigold}. We then compute two metrics:
\begin{itemize}
\item \textbf{si-MSE (scale-invariant Mean Squared Error)}~\cite{eigen2014sirmse}: Measures structural and shape-related errors between the two depth maps, invariant to global scale shifts.
\item \textbf{NMSE (Normalized Mean Squared Error)}~\cite{ohana2024well}: Quantifies the relative prediction error, complementing si-MSE for a comprehensive evaluation.
\end{itemize}
For both metrics, lower values signify better alignment with the conditioning depth map.

\subsection{Baseline Details}
Our baseline is Cosmos-Transfer1~\cite{alhaija2025cosmos-t}, which enhances a base diffusion model (Cosmos-Predict1~\cite{agarwal2025cosmos-p}) with ControlNet~\cite{zhang2023controlnet} for conditioning. For a fair comparison, we use the same base model and attach our TC-LoRA adapters. The primary architectural difference lies in the number of trainable parameters: the publicly released Cosmos-Transfer1 checkpoint includes a trainable copy of the first three transformer blocks, totaling 900M trainable parameters. In contrast, our TC-LoRA introduces only 251M trainable parameters, all contained within the shared hypernetwork, which demonstrates its memory efficiency during both the post-training process and deployment.


\begin{figure}[tp]
    \includegraphics[width=\linewidth]{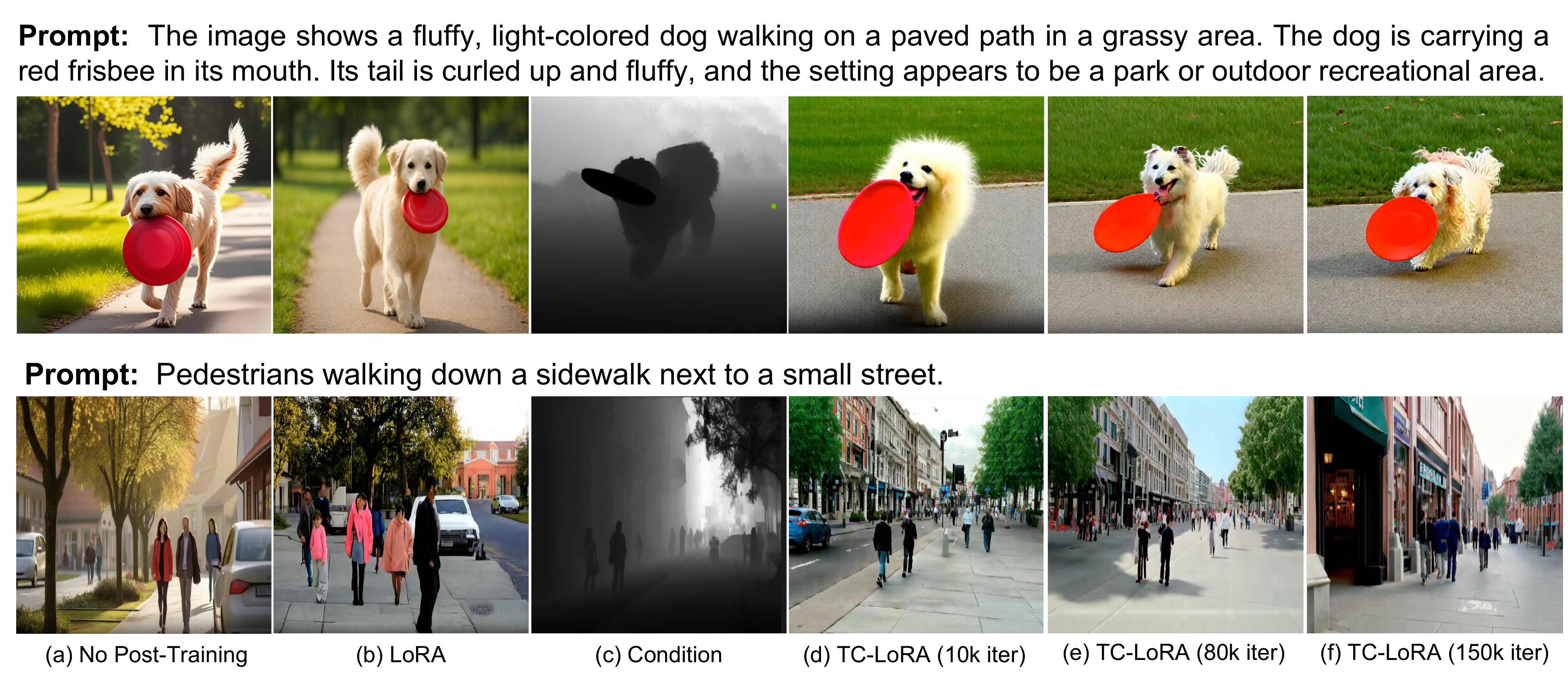}
    \caption{Effect of TC-LoRA: (a) and (b) show images generated from the given text prompt \textit{without conditioning}. With the condition input shown in (c), the generation process can be conditioned accordingly, demonstrating TC-LoRA’s role in enabling spatial correction through LoRA. Panels (d), (e), and (f) present visualization results at different TC-LoRA post-training durations. After 150k iterations, the generated image is well aligned with the condition in (c).}
    \label{fig:eval_effect}
\end{figure}

\subsection{Detailed Qualitative Analysis}
\label{appendix:C.4}
Our qualitative comparisons visually substantiate the quantitative metrics.
\begin{itemize}
\item \textbf{Baseline Comparison (Figure~\ref{fig:eval_transfer1} in the main text):} In a direct comparison across various data domains, TC-LoRA shows a superior ability to preserve details from the depth condition. For example, in the image of a dog, TC-LoRA accurately reconstructs the pose, curled tail, and the texture of the surrounding path and grass, whereas the baseline produces a structurally different dog. Similarly, in a street scene, TC-LoRA more faithfully reproduces the placement and silhouettes of pedestrians as defined by the depth map.
\item \textbf{Learning Progression (Figure~\ref{fig:eval_effect}):} This figure highlights the effectiveness of the post-training process. The unconditioned base model generates plausible but random images (a, b). When guided by the condition map (c), the model's ability to follow the structure progressively improves. At 10k iterations (d), the general composition emerges, and by 150k iterations (f), the model produces a high-fidelity image that is both semantically rich and structurally consistent with the spatial condition.
\end{itemize}

\section{On the Distinction Between Activation Conditioning and Weight Conditioning}
\label{sec:appendix_distinction}
In this section, we demonstrate why ControlNet~\cite{zhang2023controlnet} cannot be seen as an update of the weights, opposite to TC-LoRA. We remind the reader that in practice, ControlNet also takes as input the input of the diffusion model, i.e. the input noise of the diffusion process $x$. Therefore in our proof, the output of ControlNet $c$ depends on $x$.

\textbf{Proposition:} The addition of a non-constant, input-dependent vector $c(x)$ to a hidden layer activation $a_1$ cannot be expressed as a modification of a static subsequent weight matrix $W_2$.

\textit{Proof.} Let the activation of the first hidden layer be $a_1 = f(W_1x)$, where $f$ is a non-linear activation function. The pre-activation of the second layer is modified by an input-dependent vector $c(x)$, resulting in $z'_2 = W_2(a_1 + c(x))$.

Assume, for the sake of contradiction, that there exists a single, constant weight matrix $W'_2$ that is equivalent to this operation for the entire input domain of $x$. This implies the following equality must hold for all $x$:
\begin{equation}
    W'_2 a_1 = W_2(a_1 + c(x)) =W_2 a_1 + W_2 c(x)
    \label{eq:c_of_x_assumption}
\end{equation}
By rearranging, we can isolate the effect of the modification:
\begin{equation}
    (W'_2 - W_2) a_1 = W_2 c(x)
    \label{eq:c_of_x_rearranged}
\end{equation}
Let $\Delta W = W'_2 - W_2$ be the constant matrix representing the static weight update. Substituting this and the definition of $a_1$ into Equation \ref{eq:c_of_x_rearranged} yields:
\begin{equation}
    \Delta W f(W_1 x) = W_2 c(x)
    \label{eq:functional_form}
\end{equation}
This equation must hold for all $x$. Let us denote the function on the left as $L(x) = \Delta W f(W_1 x)$ and the function on the right as $R(x) = W_2 c(x)$. Our assumption that a static $W'_2$ exists is equivalent to assuming that $L(x) = R(x)$ holds for all $x$ with a constant matrix $\Delta W$.

However, the functional form of $L(x)$ is strictly determined by the architecture of the first layer (the matrix $W_1$ and the activation function $f$). In contrast, $c(x)$ can be an arbitrary function of the input, making the functional form of $R(x)$ equally arbitrary. For a general choice of $c(x)$, there is no constant matrix $\Delta W$ that can transform the function $f(W_1 x)$ into the function $W_2 c(x)$ across the entire input space. The constraint that $L(x)=R(x)$ cannot be satisfied, as the space of functions representable by $L(x)$ for a fixed $W_1$ and $f$ is a specific subset of all possible functions of $x$, while $R(x)$ can be chosen to be outside this space.

This leads to a contradiction. Therefore, the initial assumption is false, and the operation cannot be represented by a static weight update. $\square$

\subsection{TC-LoRA as a True Parametric Adaptation}
In contrast, TC-LoRA performs a true \textbf{parametric adaptation}. The weight update is generated by a hypernetwork as a direct function of time $t$ and condition $\mathbf{y}$, but it is \textbf{input-agnostic}. The adapted weight for the second layer is:
\begin{equation}
    W'_2(t, \mathbf{y}) = W_2 + \Delta W_2(t, \mathbf{y}) = W_2 + \mathbf{B}(t, \mathbf{y})\mathbf{A}(t, \mathbf{y})
\end{equation}
The output $\mathbf{y}_{weight}$ is then:
\begin{equation}
    \mathbf{y}_{weight} = f((W_2 + \mathbf{B}(t, \mathbf{y})\mathbf{A}(t, \mathbf{y})) \cdot f(W_1\mathbf{x}))
\end{equation}
Here, the weight modification is computed based on context \textit{before} it operates on the input. This is a fundamentally different mechanism: it reconfigures the layer's processing function itself. This allows TC-LoRA to learn an \textit{explicitly} time-varying strategy, enabling a deeper functional adaptation that is not achievable through input biasing alone.

\end{document}